\documentclass[]{article}
\usepackage[letterpaper]{geometry}
\usepackage{mtsummit2021}
\usepackage{times}
\usepackage{latexsym}
\usepackage[T1]{fontenc}
\usepackage[utf8]{inputenc}
\usepackage{microtype}
\usepackage{url}
\usepackage{tabularx}
\usepackage{natbib}
\usepackage{layout}
\usepackage{caption}
\usepackage{subcaption}
\usepackage{url}
\usepackage{graphicx}
\usepackage{array,multirow}
\usepackage{booktabs}
\usepackage[ruled,vlined]{algorithm2e}
\SetAlCapNameFnt{\small}
\SetAlCapFnt{\small}

\begin{document}
\title{\bf
  Active Learning for Massively Parallel Translation \\
  of Constrained Text into Low Resource Languages
}
\author{
  \name{\bf Zhong Zhou} \hfill \addr{zhongzhou@cmu.edu}\\
  \name{\bf Alex Waibel} \hfill \addr{alex@waibel.com}\\
  \addr{Language Technology Institute, School of Computer Science, Carnegie Mellon University, 5000 Forbes Ave, Pittsburgh PA 15213}
}

\maketitle

\begin{abstract}
  We translate a closed text that is
  known in advance
  and available in many 
  languages into a new and severely
  low resource language. Most human 
  translation efforts adopt 
  a portion-based approach to 
  translate consecutive
  pages/chapters in order, which 
  may not suit machine translation. We compare
  the portion-based approach that optimizes 
  coherence of the text locally with the random
  sampling approach that increases coverage 
  of the text globally. 
  Our results show that the random 
  sampling approach performs better. 
  When training on a seed corpus
  of $\sim$1,000 lines
  from the Bible and testing
  on the rest of the Bible
  ($\sim$30,000 lines), random
  sampling gives a performance
  gain of +8.5 BLEU using
  English as a simulated low resource language,
  and +1.9 BLEU using Eastern Pokomchi, a Mayan language.
  Furthermore, we 
  compare three ways of updating 
  machine translation models 
  with increasing amount of human
  post-edited data through iterations.
  We find that adding newly
  post-edited data to training after 
  vocabulary update without
  self-supervision performs the best. 
  We propose an algorithm
  for human and machine to work together
  seamlessly to translate a closed text
  into a severely low resource language.  
\end{abstract}

\section{Introduction} \label{introduction}
\begin{figure*}[t]
  \centering
  \includegraphics[width=.9\linewidth]{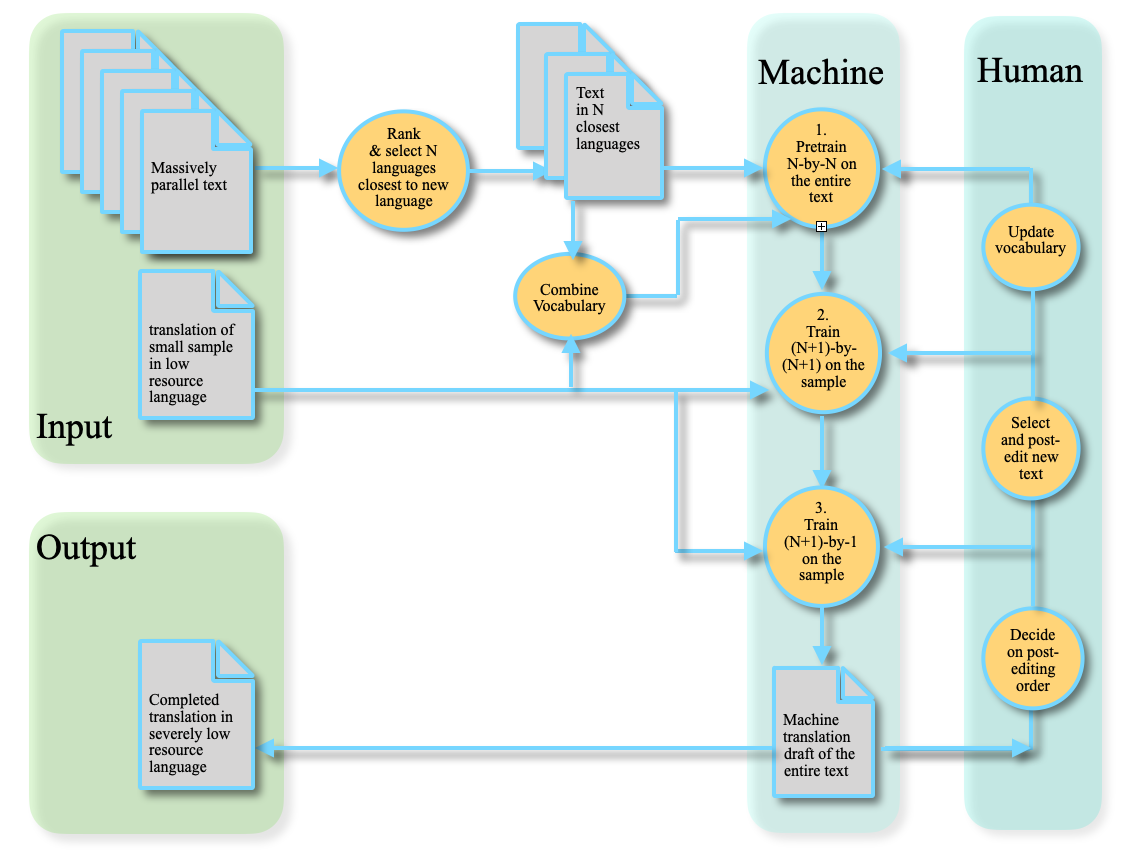}
  \caption{Proposed joint human machine translation sequence for a given closed text. }
  \label{fig:hmt_algo}
\end{figure*}

Machine translation has flourished ever since
the first computer was made
\citep{hirschberg2015advances, popel2020transforming}.
Over the years, human translation is assisted by machine
translation to 
remove human bias and translation 
capacity limitations
\citep{koehn2009interactive, li2014comparison, savoldi2021gender, bowker2002computer, bowker2010computer, koehn2009process}.
By learning human translation taxonomy
and post-editing styles,
machine translation borrows many ideas
from human translation to improve performance
through active learning 
\citep{settles2012active, carl2011taxonomy, denkowski2015machine}.
We propose a workflow to bring human
translation and machine translation to work together
seamlessly in translation of a closed text
into a severely low resource language as shown in
Figure \ref{fig:hmt_algo} and Algorithm \ref{algo:proposedtrans}.
 
Given a closed text that has many existing
translations in different languages, we are
interested in translating it into a severely
low resource language well.
Researchers recently have shown achievements
in translation using
very small seed parallel
corpora in low resource languages
\citep{lin2020pre, qi2018and, zhong2018massively}.
Construction methods of such seed corpora 
are therefore pivotal in translation performance. 
Historically, this is 
mostly determined by field linguists'
experiential and intuitive discretion. 
Many human translators
employ a portion-based strategy when translating large texts.
For example, translation of the book ``The Little Prince''
may be divided into smaller tasks
of translating 27 chapters, or
even smaller translation units like a few consecutive pages.
Each translation unit contains
consecutive sentences.
Consequently, machine translation often uses
seed corpora that are chosen based on human
translators' preferences, but may not be optimal
for machine translation. 

\begin{algorithm*}[t]
\small
 \KwIn{A text of $N$ lines consisting multiple books/portions, parallel in $L$ source languages}  
 \KwOut{A full translation in the target low resource language, $l'$}
  0. Initialize translation size, $n = 0$, vocabulary size, $v = 0$, vocabulary update size, $\triangle v = 0$ \;
  1. Randomly sample $S$ ($\sim$1,000) sentences with vocabulary size $v_S$ for human translators to produce the seed corpus, update $n = S$, $v = v_S$ \;
  2. Rank and pick a family of close-by languages by linguistic, distortion or performance metric \;
  \While{$n < N$} {
    \If{$\triangle v > 0 $ } {
  3. Pretrain on the full texts of neighboring languages \;
  }
  4. Train on the $n$ sentences of all languages in multi-source multi-target configuration \;
  5. Train on the $n$ sentences of all languages in multi-source single-target configuration \;
  6. Combine translations from all source languages using the centeredness measure \; 
  7. Review all books/portions of the translation draft \;
  8. Pick a book/portion with $n'$ lines and $v'$ more vocabulary \;
  9. Complete human post-editing of the portion chosen, $v = v + v'$, $n = n + n'$, $\triangle v = v'$ \;
  }
 return full translation co-produced by human (Step 1, 7-9) and machine (Step 0, 2-6) translation \;
  \caption{Proposed joint human machine translation sequence for a given closed text.}
  \label{algo:proposedtrans}
\end{algorithm*}

\begin{table*}[t]
  \centering
  \small
  \resizebox{\textwidth}{!}{
    \begin{tabularx}{\textwidth}{p{3.4cm}p{3.2cm}p{0.9cm}p{0.9cm}p{0.9cm}p{0.9cm}}
    \toprule
    Book & Author & Books & Chapters & Pages & Languages\\    
    \midrule
    The Bible & Multiple & 66 & 1,189 & 1,281 & 689 \\
    The Little Prince & Antoine de Saint Exupéry & 1 & 27 & 96 & 382 \\
    Dao De Jing & Laozi & 1 & 81 & $\sim$10 & $>$250 \\
    COVID-19 Wiki Page & Multiple & 1 & 1 & $\sim$50 & 155 \\
    The Alchemist & Paulo Coelho & 1 & 2 & 163 & 70\\
    Harry Potter & J. K. Rowling & 7 & 199 & 3,407 & 60\\
    The Lord of the Rings & J. R. R. Tolkien & 6 & 62 & 1,037 & 57\\
    Frozen Movie Script & Jennifer Lee & 1 & 112 & $\sim$40 & 41\\
    The Hand Washing Song & Multiple & 1 & 1 & 1 & 28\\
    Dream of the Red Chamber & Xueqin Cao & 2 & 120 & 2500 & 23\\
    Les Misérables & Victor Hugo & 68 & 365 & 1,462 & 21\\
    \bottomrule
  \end{tabularx} }
  \caption{Examples of different texts with the
    number of languages translated to date \citep{unesco1932world, mayer2014creating, de2019principito, zi1993dao, covid19wiki, coelho2014alchemist, rowling2019harry, tolkien2012lord, jennifer2013frozen, thampi2020s, xueqin2011dream, hugo1863miserables}.}
    \label{table:exbooks}
\end{table*}

We propose to use a random sampling approach
to build seed corpora when resources are extremely
limited. In other words, when field linguists have
limited time and resources, which lines would be
given priority? Given a closed text, we propose
that it would beneficial if field linguists translate
randomly sampled $\sim$1,000 lines first, getting the first
machine translated draft of the whole text, and then
post-edit to obtain final translation of each portion
iteratively as shown in Algorithm \ref{algo:proposedtrans}.
We recognize that the portion-based translation is very helpful in
producing quality translation with
formality, cohesion and contextual relevance.
Thus, our proposed way is not to
replace the portion-based approach, but instead,
to get the best of both worlds and
to expedite the translation process as shown in
Figure \ref{fig:hmt_algo}. 

The main difference of the two approaches is that
the portion-based approach focuses on preserving
coherence of the text locally, while the random-sampling
approach focuses on increasing coverage of the text
globally. Our results show that the random
sampling approach performs better. 
When training on a seed corpus of $\sim$1,000 lines
from the Bible and testing on the rest of the Bible
($\sim$30,000 lines), random sampling beats
the portion-based approach by +8.5 BLEU using
English as a simulated low resource language training on
a family of languages built on the distortion measure, and by
+1.9 using a Mayan language, Eastern Pokomchi, training
on a family of languages based on the linguistic definition.
Using random sampling, machine translation
is able to produce a high-quality first draft of the whole
text that expedites the subsequent iterations of translation efforts.

Moreover, we compare three different ways of incorporating
incremental post-edited data during
the translation process. We find that
self-supervision using the whole translation draft
affects performance 
adversely, and is best to be avoided. We also show that adding
the newly post-edited text to
training with vocabulary update performs the best. 

\section{Related Works} \label{relatedwork}
\subsection{Human Translation and Machine Translation}
Machine translation began about the same time as the first
computer \citep{hirschberg2015advances, popel2020transforming}.
Over the years,
human translators have different 
reactions to machine translation advances,
mixed with doubt or fear \citep{hutchins2001machine}.
Some researchers study human translation taxonomy
for machine to better assist human
translation and post-editing efforts \citep{carl2011taxonomy, denkowski2015machine}.
Human translators benefit from machine assistance
as human individual bias and translation capacity limitations
are compensated for by large-scale
machine translation
\citep{koehn2009interactive, li2014comparison, savoldi2021gender, bowker2002computer, bowker2010computer, koehn2009process}.
On the other hand, machine translation 
benefits from professional human translators'
context-relevant and culturally-appropriate 
translation and post-editing efforts \citep{hutchins2001machine}.
Severely low resource translation is a fitting ground for 
close human machine collaboration
\citep{zong2018research, carl2011taxonomy, martinez2003human}.

\subsection{Severely Low Resource Text-based Translation}
Many use multiple rich-resource
languages to translate to a low resource language
using multilingual methods 
\citep{johnson2017google, ha2016toward, firat2016multi, zoph2016multi, zoph2016transfer, adams2017cross, gillick2016multilingual, zhong2018massively, zhou2018paraphrases}.
Some use data selection for active learning 
\citep{eck2005low}. Some use
as few as $\sim$4,000 lines \citep{lin2020pre, qi2018and}
and $\sim$1,000 lines \citep{zhou2021family} of data.
Some do not use low resource data \citep{neubig2018rapid, karakanta2018neural}. 

\subsection{Active Learning and Random Sampling}
Active learning has long been used
in machine translation
\citep{settles2012active, ambati2012active, eck2005low, haffari2009active, gonzalez2012active, miura2016selecting, gangadharaiah2009active}.   
Random sampling and data selection
has been successful
\citep{kendall1938randomness, knuth19913, clarkson1989applications, sennrich2015improving, hoang2018iterative, he2016dual, gu2018meta}.
The mathematician Donald Knuth uses the population
of Menlo Park to illustrate the value of random sampling
\citep{knuth19913}.

\begin{table*}[t]
  \centering
  \small
  \resizebox{\textwidth}{!}{
  \begin{tabular}{p{1.1cm}p{0.4cm}p{0.65cm}|p{0.4cm}p{0.6cm}|
      p{1.1cm}p{0.4cm}p{0.65cm}|p{0.4cm}p{0.6cm}|
      p{1.1cm}p{0.4cm}p{0.65cm}|p{0.4cm}p{0.6cm}}
    \toprule
    \multicolumn{15}{c}{Input Language Family} \\
    \midrule
    \multicolumn{5}{c|}{By Linguistics} & \multicolumn{5}{c|}{By Distortion} & \multicolumn{5}{c}{By Performance} \\
    \midrule
    \multicolumn{5}{c|}{\textit{FAMO$^+$}} & \multicolumn{5}{c|}{\textit{FAMD}} & \multicolumn{5}{c}{\textit{FAMP}} \\
    \midrule
    Training & \multicolumn{2}{c|}{\textit{Luke}} & \multicolumn{2}{c|}{\textit{Rand}} &
    Training & \multicolumn{2}{c|}{\textit{Luke}} & \multicolumn{2}{c|}{\textit{Rand}} &
    Training & \multicolumn{2}{c|}{\textit{Luke}} & \multicolumn{2}{c}{\textit{Rand}} \\
    \midrule
    Testing & \textit{Best} & \textit{All} & \textit{Best} & \textit{All} &
    Testing & \textit{Best} & \textit{All} & \textit{Best} & \textit{All} &
    Testing & \textit{Best} & \textit{All} & \textit{Best} & \textit{All} \\
    \midrule
    Combined & 37.9 & 21.9 & 42.8 & 28.6 &
    Combined & 38.6 & 22.9 & 44.8 & 31.4 &
    Combined & 40.2 & 23.7 & 44.6 & 30.6 \\
    \midrule
    German & 35.6 & 20.0 & 40.8 & 26.5 &
    German & 37.0 & 20.8 & 42.7 & 28.8 &
    German & 38.0 & 21.3 & 41.6 & 28.2\\
    Danish & 36.7 & 19.0 & 38.2 & 25.9 &
    Danish & 37.3 & 19.6 & 39.5 & 28.0 &
    Danish & 38.4 & 19.9 & 39.2 & 27.5 \\
    Dutch & 36.4 & 20.4 & 39.7 & 27.2 &
    Dutch & 36.4 & 21.1 & 41.9 & 29.6 &
    Dutch & 37.5 & 21.6 & 41.6 & 28.9 \\
    Norwegian & 36.5 & 20.2 & 40.0 & 26.9 &
    Norwegian & 37.2 & 20.8 & 41.4 & 29.1 &
    Norwegian & 37.5 & 21.1 & 41.0 & 28.4 \\
    Swedish & 34.9 & 19.7 & 39.9 & 26.2 &
    Afrikaans & 38.3 & 22.2 & 42.8 & 30.5 &
    Afrikaans & 39.5 & 22.9 & 42.3 & 29.8 \\
    Spanish & 36.8 & 21.5 & 39.8 & 27.6 &
    Marshallese & 35.1 & 21.6 & 41.4 & 28.8 &
    Spanish & 38.7 & 22.9 & 41.9 & 29.0 \\
    French & 36.0 & 19.7 & 39.6 & 26.1 &
    French & 36.2 & 20.3 & 41.1 & 28.3 &
    French & 37.3 & 20.7 & 40.5 & 27.5 \\
    Italian & 36.7 & 20.6 & 38.4 & 26.9 &
    Italian & 37.3 & 21.0 & 40.6 & 29.1 &
    Italian & 38.6 & 21.8 & 39.9 & 28.5 \\
    Portuguese & 32.4 & 15.8 & 30.1 & 21.3 &
    Portuguese & 33.2 & 16.5 & 33.6 & 24.0 &
    Portuguese & 33.7 & 16.3 & 33.1 & 22.9 \\
    Romanian & 34.9 & 19.3 & 37.1 & 26.0 &
    Frisian & 36.4 & 21.6 & 43.0 & 29.8 &
    Frisian & 37.8 & 22.3 & 42.2 & 29.1 \\
    \bottomrule
  \end{tabular}
  }
  \caption{Performance training on 1,093 lines of Eastern Pokomchi data on \textit{FAMO$^+$}, \textit{FAMD} and \textit{FAMP}. We train using the portion-based approach in \textit{Luke}, and using random sampling in \textit{Rand}. During testing, \textit{Best} is the book with highest BLEU score, and \textit{All} is the performance on $\sim$29,000 lines of test data \textsuperscript{\ref{fn1}}.}
  \label{table:enRandCompare}
\end{table*}
\begin{table*}[t] 
  \centering
  \small
  \resizebox{\textwidth}{!}{
  \begin{tabular}{p{1.1cm}p{0.4cm}p{0.65cm}|p{0.4cm}p{0.6cm}|
  p{1.1cm}p{0.4cm}p{0.65cm}|p{0.4cm}p{0.6cm}|
  p{1.1cm}p{0.4cm}p{0.65cm}|p{0.4cm}p{0.6cm}}
    \toprule
    \multicolumn{15}{c}{Input Language Family} \\
    \midrule
    \multicolumn{5}{c|}{By Linguistics} & \multicolumn{5}{c|}{By Distortion} & \multicolumn{5}{c}{By Performance} \\
    \midrule
    \multicolumn{5}{c|}{\textit{FAMO$^+$}} & \multicolumn{5}{c|}{\textit{FAMD}} & \multicolumn{5}{c}{\textit{FAMP}} \\
    \midrule
    Training & \multicolumn{2}{c|}{\textit{Luke}} & \multicolumn{2}{c|}{\textit{Rand}} &
    Training & \multicolumn{2}{c|}{\textit{Luke}} & \multicolumn{2}{c|}{\textit{Rand}} &
    Training & \multicolumn{2}{c|}{\textit{Luke}} & \multicolumn{2}{c}{\textit{Rand}} \\
    \midrule       
    Testing & \textit{Best} & \textit{All} & \textit{Best} & \textit{All} &
    Testing & \textit{Best} & \textit{All} & \textit{Best} & \textit{All} &
    Testing & \textit{Best} & \textit{All} & \textit{Best} & \textit{All} \\
    \midrule
    Combined & 23.1 & 8.6 & 19.7 & 10.5 & 
    Combined & 23.3 & 8.5 & 17.7 & 9.5 & 
    Combined & 22.4 & 7.2 & 15.8 & 7.8 \\
    \midrule
    Chuj & 21.8 & 7.9 & 16.5 & 9.8 &
    Chuj & 22.0 & 7.9 & 15.4 & 8.9 & 
    Chuj & 21.8 & 7.0 & 13.2 & 7.3 \\ 
    Cakchiquel & 22.3 & 7.9 & 18.2 & 9.9 &
    Cakchiquel & 22.4 & 7.9 & 17.3 & 9.1 &
    Cakchiquel & 21.2 & 6.9 & 14.8 & 7.4 \\ 
    Guajajara & 19.9 & 7.1 & 14.7 & 8.9 &
    Guajajara & 19.2 & 6.9 & 14.2 & 8.2 &
    Guajajara & 18.9 & 5.9 & 10.6 & 6.6 \\
    Mam & 22.2 & 8.6 & 19.7 & 10.6 &
    Russian & 22.2 & 7.3 & 13.7 & 8.5 &
    Mam & 21.9 & 7.5 & 17.1 & 8.0 \\ 
    Kanjobal & 21.8 & 8.1 & 17.5 & 10.0 &
    Toba & 22.0 & 8.3 & 16.8 & 9.4 &
    Kanjobal & 21.6 & 7.1 & 13.8 & 7.6 \\ 
    Cuzco & 22.4 & 7.8 & 17.7 & 9.8 &
    Myanmar & 19.2 & 5.3 & 10.7 & 6.5 &
    Thai & 21.9 & 6.3 & 10.5 & 7.0 \\
    Ayacucho & 21.6 & 7.6 & 18.5 & 9.7 &
    Slovenský & 22.2 & 7.5 & 13.5 & 8.7 &
    Dadibi & 19.9 & 6.2 & 15.3 & 6.9 \\
    Bolivian & 22.3 & 7.8 & 17.4 & 9.8 &
    Latin & 22.0 & 7.8 & 14.8 & 9.0 &
    Gumatj & 19.2 & 3.8 & 8.9 & 3.3 \\
    Huallaga & 22.2 & 7.7 & 18.0 & 9.7 &
    Ilokano & 22.6 & 8.4 & 17.8 & 9.4 &
    Navajo & 21.4 & 6.5 & 13.5 & 7.3 \\
    Aymara & 21.5 & 7.5 & 18.6 & 9.6 &
    Norwegian & 22.6 & 8.3 & 16.7 & 9.4 &
    Kim & 21.6 & 7.0 & 13.9 & 7.5 \\
    \bottomrule
  \end{tabular}
  }
  \caption{Performance training on 1,086 lines of Eastern Pokomchi data on \textit{FAMO$^+$}, \textit{FAMD} and \textit{FAMP}. We train using the portion-based approach in \textit{Luke}, and using random sampling in \textit{Rand}. During testing, \textit{Best} is the book with highest BLEU score, and \textit{All} is the performance on $\sim$29,000 lines of test data \textsuperscript{\ref{fn1}}.}
    \label{table:phRandCompare}
\end{table*}

\begin{table}[t]
  \centering
  \small
  \begin{tabularx}{\columnwidth}{p{1.8cm}p{1.9cm}p{2.7cm}p{2.3cm}p{2.3cm}}
    \toprule
    Source & \textit{Seed} & \textit{Self-Supervised} & \textit{Old-Vocab} & \textit{Updated-Vocab} \\
    \midrule
    Combined & 30.8 & 24.4 (-6.4) & 32.1 (+1.3) & 32.4 (+1.6) \\
    \midrule
    Danish & 27.7 & 21.6 (-6.1) & 28.8 (+1.1) & 29.2 (+1.5) \\
    Norwegian & 28.6 & 22.5 (-6.1) & 29.8 (+1.2) & 30.2 (+1.6) \\
    Italian & 28.7 & 22.3 (-6.4) & 29.8 (+1.1) & 30.2 (+1.5) \\
    Afrikaans & 30.1 & 23.8 (-6.3) & 31.4 (+1.3) & 31.6 (+1.5) \\
    Dutch & 29.2 & 22.9 (-6.3) & 30.3 (+1.1) & 30.6 (+1.4) \\
    Portuguese & 23.8 & 18.3 (-5.5) & 24.6 (+0.8) & 25.0 (+1.2) \\
    French & 27.8 & 21.7 (-6.1) & 28.9 (+1.1) & 29.4 (+1.6) \\
    German & 28.4 & 22.4 (-6.0) & 29.5 (+1.1) & 29.9 (+1.5) \\
    Marshallese & 28.4 & 22.4 (-6.0) & 29.5 (+1.1) & 29.9 (+1.5) \\
    Frisian & 29.3 & 23.2 (-6.1) & 30.4 (+1.1) & 30.8 (+1.5) \\
    \bottomrule
    \end{tabularx}
  \caption{Comparing three ways of adding the newly post-edited book of 1 Chronicles \textsuperscript{\ref{fn1}}. \textit{Seed} is the baseline of training on the seed corpus alone, \textit{Old-Vocab} skips the vocabulary update while \textit{Updated-Vocab} has vocabulary update. \textit{Self-Supervised} adds the complete translation draft in addition to the new book. 
  }
  \label{table:1chron}
\end{table}

\section{Methodology}\label{method}
We train our models using a state-of-the-art multilingual transformer
by adding language labels to each source sentence
\citep{johnson2017google, ha2016toward, zhong2018massively, zhou2018paraphrases}.
We borrow the order-preserving named entity translation
method by replacing each named entity with \texttt{\_\_NE}s
\citep{zhou2018paraphrases} using a multilingual
lexicon table that covers 124 source languages and 2,939
named entities \citep{zhou2021family}. For example, the sentence ``Somchai calls Juan''
is transformed to ``\texttt{\_\_opt\_src\_en \_\_opt\_tgt\_ca} \texttt{\_\_NE0} calls \texttt{\_\_NE1}''
to translate to Chuj. 
We use families of close-by languages constructed by
ranking 124 source languages by distortion measure (\textit{FAMD}),
performance measure (\textit{FAMP})
and linguistic family (\textit{FAMO$^+$});
the distortion measure ranks languages by decreasing probability
of zero distortion, while the performance measure incorporates
an additional probability of fertility equalling one \citep{zhou2021family}.
Using families constructed, we pretrain our model first 
on the whole text of nearby languages,
then we train on the $\sim$1,000 lines
of low resource data and the corresponding lines in other languages 
in a multi-source multi-target fashion.
We finally train on the $\sim$1,000 lines in a multi-source single-target
fashion \citep{zhou2021family}.

We combine translations
of all source languages into one. Let all $N$ translations
be $t_i, i = 1, \ldots, N$ and let similarity
between translations $t_i$ and $t_j$ be $S_{ij}$. 
We rank all translations according to how
centered it is with respect to other sentences by
summing all its similarities to the rest through $\sum_j S_{ij}$ for $i = 1, \ldots, N$.
We take the most centered translation
for every sentence, $\max_i \sum_j S_{ij}$, to build the combined
translation output. The expectation of the combined score is higher than that
of any of the source languages \citep{zhou2021family}.  

Our work differs from the past research in that we put
low resource translation into the broad collaborative
scheme of human machine translation. 
We compare the portion-based approach with 
the random sampling approach in building seed corpora.
We also compare three methods of updating models with
increasing amount of human post-edited data.
We add the newly post-edited data to training in three ways: 
with vocabulary update, 
without vocabulary update, 
or incorporating the whole translation draft in a
self-supervised fashion additionally. 
For best performance,
we build the seed corpus by random sampling, 
update vocabulary iteratively, 
and add newly post-edited data to 
training without self-supervision.
We also have a larger test set,
we test on $\sim$30,000 lines rather than
$\sim$678 lines from existing research \textsuperscript{\ref{fn1}}. 

We propose a joint human machine translation
workflow in Algorithm \ref{algo:proposedtrans}.
After pretraining on neighboring languages in Step 3,
we iteratively train on the randomly sampled
seed corpus of low resource data in Step 4 and 5.
The reason we include both Step 4 and 5 in our algorithm
is because training both steps iteratively performs 
better than training either one \citep{zhou2021family}. 
Our model produces a translation draft of the whole text. 
Since the portion-based approach has the advantage
with formality, cohesion and contextual relevance, human
translators may pick and post-edit portion-by-portion iteratively. 
The newly post-edited data with updated vocabulary is added to
the machine translation models without self-supervision. 
In this way, machine translation systems
rely on quality parallel corpora that are incrementally produced
by human translators.
Human translators lean on machine
translation for quality translation draft to expedite translation.
This creates a synergistic collaboration between
human and machine. 

\section{Data}\label{data}
We work on the Bible in 124 source languages \citep{mayer2014creating}, and
have experiments for English, a simulated
language, and Eastern Pokomchi, a Mayan language.
We train on $\sim$1,000 lines of low resource data and on full texts for all
the other languages.
We aim to translate the
rest of the text ($\sim$30,000 lines) into the low resource language.
In pretraining, we use
80\%, 10\%, 10\% split for training, validation
and testing. In training, we
use 3.3\%, 0.2\%, 96.5\% split for training, validation
and testing. Our test size is >29 times of the training size \textsuperscript{\ref{fn1}}.
We use the book "Luke" for the portion-based approach as suggested by many 
human translators. 

Training on $\sim$100 million parameters with Geforce RTX 2080 Ti,
we employ a 6-layer encoder and a 6-layer decoder with
512 hidden states, 8 attention heads,
512 word vector size, 2,048 hidden units,
6,000 batch size, 0.1 label smoothing,
2.5 learning rate, 0.1 dropout and attention dropout,
an early stopping patience of 5 after 190,000 steps,
``BLEU'' validation metric,
``adam'' optimizer and ``noam'' decay method \citep{klein2017opennmt, papineni2002bleu}. We increase
patience to 25 for larger data in the second stage of training in Figure
\ref{fig:curve_en_hmt} and \ref{fig:curve_ph_hmt}.

\section{Results}\label{results}
We observe that random sampling performs 
better than the portion-based approach.
In Table \ref{table:enRandCompare} and \ref{table:phRandCompare},
random sampling gives a performance gain of +8.5
for English on FAMD and +1.9 for Eastern Pokomchi on FAMO$^+$ \footnote{
  \label{fn1}
  Previously, we test on $\sim$30,000 lines excluding
  the $\sim$1,000 lines of training and validation data. In this version of our paper,
  we test on the intersection of different test sets. 
  In Table \ref{table:enRandCompare} and \ref{table:phRandCompare}, we test
  on $\sim$29,000 lines of data of the Bible excluding both the book of Luke
  and the randomly sampled $\sim$1,000 lines. In Table~\ref{table:1chron}, we
  evaluate on $\sim$29,000 lines of data of the Bible excluding both 
  the randomly sampled $\sim$1,000 lines and the book of 1 Chronicles. 
}.
The performance gain for
Eastern Pokomchi may be lower because Mayan languages are
morphologically rich, complex, isolated and opaque
\citep{aissen2017mayan, clemens2015ergativity, england2011grammar}.
English is closely related to many languages due to
colonization and globalization even though it is
artificially constrained in size \citep{bird2020decolonising}.
This may explain why Eastern Pokomchi
benefits less. 

To simulate human translation efforts in Step 7 and 8 in Algorithm
\ref{algo:proposedtrans}, we rank 66 books of the Bible
by BLEU scores on English's FAMD and Eastern
Pokomchi's FAMO$^+$. We assume that BLEU ranking is available
to us to simulate human judgment.
In reality, this step is realized by human translators
skimming through the translation draft and comparing performances 
of different books by intuition and experience.
In Section \ref{conclusion},
we will discuss the limitation of this assumption.
Performance ranking of the simulated
low resource language may differ from that of the
actual low resource language.
But the top few may coincide
because of the nature of the text, independent of the language.
In our results, we observe that narrative books performs better than
philosophical or poetic books. The book of 1 Chronicles
performs best for both English and Eastern Pokomchi with random sampling.
A possible explanation is
that the book of 1 Chronicles is mainly narrative,
and contains many named
entities that are translated well by the
order-preserving lexiconized model.
We included the BLEU scores of the
best-performing book in Table \ref{table:enRandCompare} and \ref{table:phRandCompare}. 
Note that only scores of ``All'' are comparable across experiments
trained on the book of Luke with those trained by
random sampling as they evaluate on the same set \textsuperscript{\ref{fn1}}. 
For the best-performing book, it is the book of 1 Chronicles for random sampling,
and the book of Mark or the book of Matthew for experiments trained on the book of Luke.
Thus, we cannot compare BLEU scores for the best-performing books across experiments. 
We include them in the tables to show the quality of the translation draft human translators will
work on if they proceed to translate the best-performing book. 

In Table \ref{table:1chron}, we compare
three different ways of
updating the machine translation models by
adding a newly post-edited book that
human translators produced.
We call the baseline without addition of the new book \textit{Seed}.
\textit{Updated-Vocab} adds the new book to 
training with updated vocabulary
while \textit{Old-Vocab} skips the vocabulary update.
\textit{Self-Supervised} adds the whole translation draft
of $\sim$30,000 lines to pretraining in addition to the new
book. Self-supervision refers to using the small seed corpus to translate
the rest of the text which is subsequently used to train the model. 
We observe that the \textit{Self-Supervised} performs the
worst among the three. Indeed, \textit{Self-Supervised} performs even
worse than the baseline \textit{Seed}.
This shows that quality is much more important than
quantity in severely low resource translation. It is better
for us not to add the whole translation draft to the
pretraining as it affects performance adversely.

On the other hand, we see that both \textit{Updated-Vocab} and
\textit{Old-Vocab} performs better than \textit{Seed} and
\textit{Self-Supervised}.
\textit{Updated-Vocab}'s performance is better than
\textit{Old-Vocab}. An explanation could be that \textit{Updated-Vocab}
has more expressive power with updated vocabulary.
Therefore, in our proposed algorithm,
we prefers vocabulary update in each iteration. If
the vocabulary has not increased, we may skip 
pretraining to expedite the process. 

\begin{figure*}
\centering
\begin{subfigure}{.45\textwidth}
  \centering
  \includegraphics[width=\linewidth]{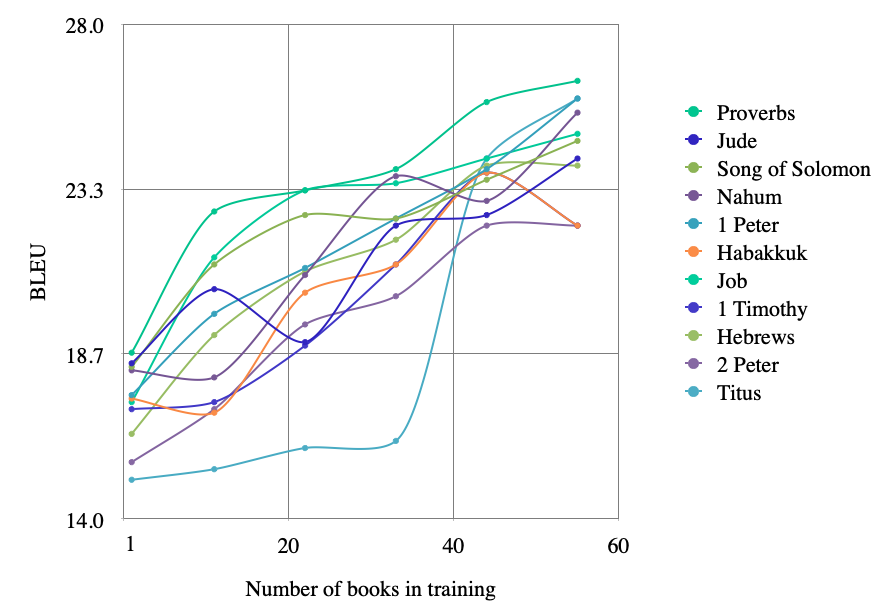}
  \caption{English}
  \label{fig:curve_en_hmt}
\end{subfigure}
\begin{subfigure}{.45\textwidth}
  \centering
  \includegraphics[width=\linewidth]{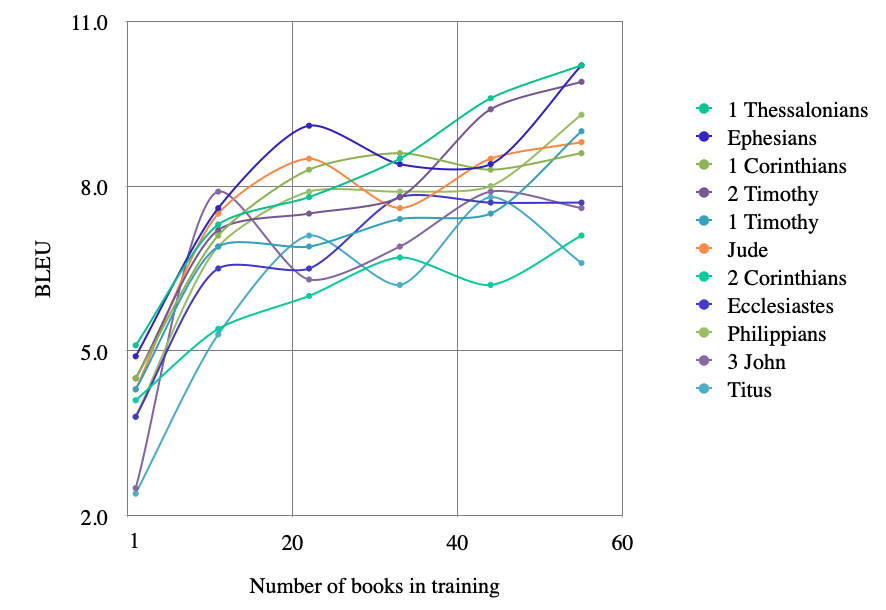}
  \caption{Eastern Pokomchi}
  \label{fig:curve_ph_hmt}
\end{subfigure}
\caption{Performance of the most difficult 11 books with increasing number of training books.}
\end{figure*}

We show how the algorithm is put into practice for English and Eastern Pokomchi 
in Figure \ref{fig:curve_en_hmt} and \ref{fig:curve_ph_hmt}.
We take the worst-performing 11 books as the held-out test set,
and divide the other 55 books of the Bible into 5 portions. 
Each portion contains 11 books. We translate
the text by using the randomly sampled $\sim$1,000 lines of seed corpus
first, and then proceed with human machine translation
in Algorithm \ref{algo:proposedtrans}
in 5 iterations with increasing number of post-edited portions.

For English, we observe that philosophical books like ``Proverbs'' and poetry books like
``Song of Solomon'' perform very badly in the beginning, but begin to achieve
above 20 BLEU scores after adding 11 books of training data. 
This reinforces our earlier result that $\sim$20\% of the text
is sufficient for achieving high-quality translation \citep{zhong2018massively}.  
However, some books like ``Titus'' remains
difficult to translate even after adding 33 books of training data. This shows
that adding data may benefit some books more than the others.
A possible explanation is that there are
multiple authors of the Bible, and books differ from each other
in style and content. Some books are closely related to each other, and may
benefit from translations of other books. But some may be
very different and benefit much less. 

For Eastern Pokomchi, though the performance
of the most difficult 11 books never reach
BLEU score of 20s like that of English experiments,
all books have BLEU scores that are steadily
increasing. Challenges remain for Eastern Pokomchi,
a Resource 0 language \citep{joshi2020state}. We hope to
work with native Mayan speakers to see ways we may improve the
results. 

\section{Conclusion}\label{conclusion}
We propose to use random sampling to build
seed parallel corpora instead of using
the portion-based approach
in severely low resource settings. Training on $\sim$1,000 lines, 
the random sampling approach outperforms 
the portion-based approach by +8.5
for English's FAMD, and by +1.9
for Eastern Pokomchi's FAMO$^+$.
We also compare three different ways
of updating the machine translation models by
adding newly post-edited data iteratively.
We find that vocabulary update is necessary, but 
self-supervision by pretraining with whole translation draft 
is best to be avoided. 

One limitation of our work is that in real life scenarios, we do not have the reference
text in low resource languages to 
produce the BLEU scores to decide the 
post-editing order. Consequently, field linguists
need to skim through and decide the post-editing order 
based on intuition. However, computational models
can still help. One potential way to tackle it 
is that we can train on $\sim$1,000 lines from
another language with available text and
test on the 66 books. Since our results show that
the literary genre plays important role in the performance ranking,
it would be reasonable to determine the order
using a ``held-out language'' and then using that
to determine order in the target low resource language. 
In the future, we
would like to work with human translators who understand and speak 
low resource languages. 

Another concern human translators may have is the creation
of randomly sampled seed corpora. To gauge the amount of interest
or inertia, we have interviewed some
human translators and many are interested. However,
it is unclear whether human translation quality of randomly sampled data
differs from that of the traditional portion-based approach. 
We hope to work with human translators closely to determine whether the translation 
quality difference is manageable. 

We are also curious how our model will perform with
large literary works like ``Lord of the Rings'' and "Les Misérables". 
We would like to see whether it will translate well
with philosophical depth and literary complexity. However,
these books often have copyright
issues and are not as easily available as the Bible data.
We are interested in collaboration with teams who have multilingual
data for large texts, especially multilingual COVID-19 data. 

\pagebreak
\bibliographystyle{apalike}
\bibliography{thesis}

\end{document}